\documentclass[runningheads]{llncs}
\usepackage{graphicx}
\usepackage{hyperref}
\hypersetup{
     colorlinks   = true,
     citecolor    = blue
}
\usepackage{array}
\newcolumntype{P}[1]{>{\centering\arraybackslash}p{#1}}

\usepackage{tikz}
\usepackage{comment}
\usepackage{amsmath,amssymb} 
\usepackage{color}
\usepackage{cite}

\begin{document}
\pagestyle{headings}
\mainmatter
\def\ECCVSubNumber{100}  

\title{AttnGrounder: Talking to Cars with Attention} 

\titlerunning{AttnGrounder}
%
\author{Vivek Mittal}
\authorrunning{V. Mittal}
%
\institute{Indian Institute of Technology, Mandi, India \\
\email{b18153@students.iitmandi.ac.in}}
\maketitle

\begin{abstract}
We propose Attention Grounder (AttnGrounder), a single-stage end-to-end trainable model for the task of visual grounding. Visual grounding aims to localize a specific object in an image based on a given natural language text query.  Unlike previous methods that use the same text representation for every image region, we use a visual-text attention module that relates each word in the given query with every region in the corresponding image for constructing a region dependent text representation. Furthermore, for improving the localization ability of our model, we use our visual-text attention module to generate an attention mask around the referred object. The attention mask is trained as an auxiliary task using a rectangular mask generated with the provided ground-truth coordinates. We evaluate AttnGrounder on the Talk2Car dataset and show an improvement of 3.26\% over the existing methods. Code is available at \url{https://github.com/i-m-vivek/AttnGrounder}.

\end{abstract}

\section{Introduction}

\begin{figure}[t]
\centering
\includegraphics[height=3.6cm]{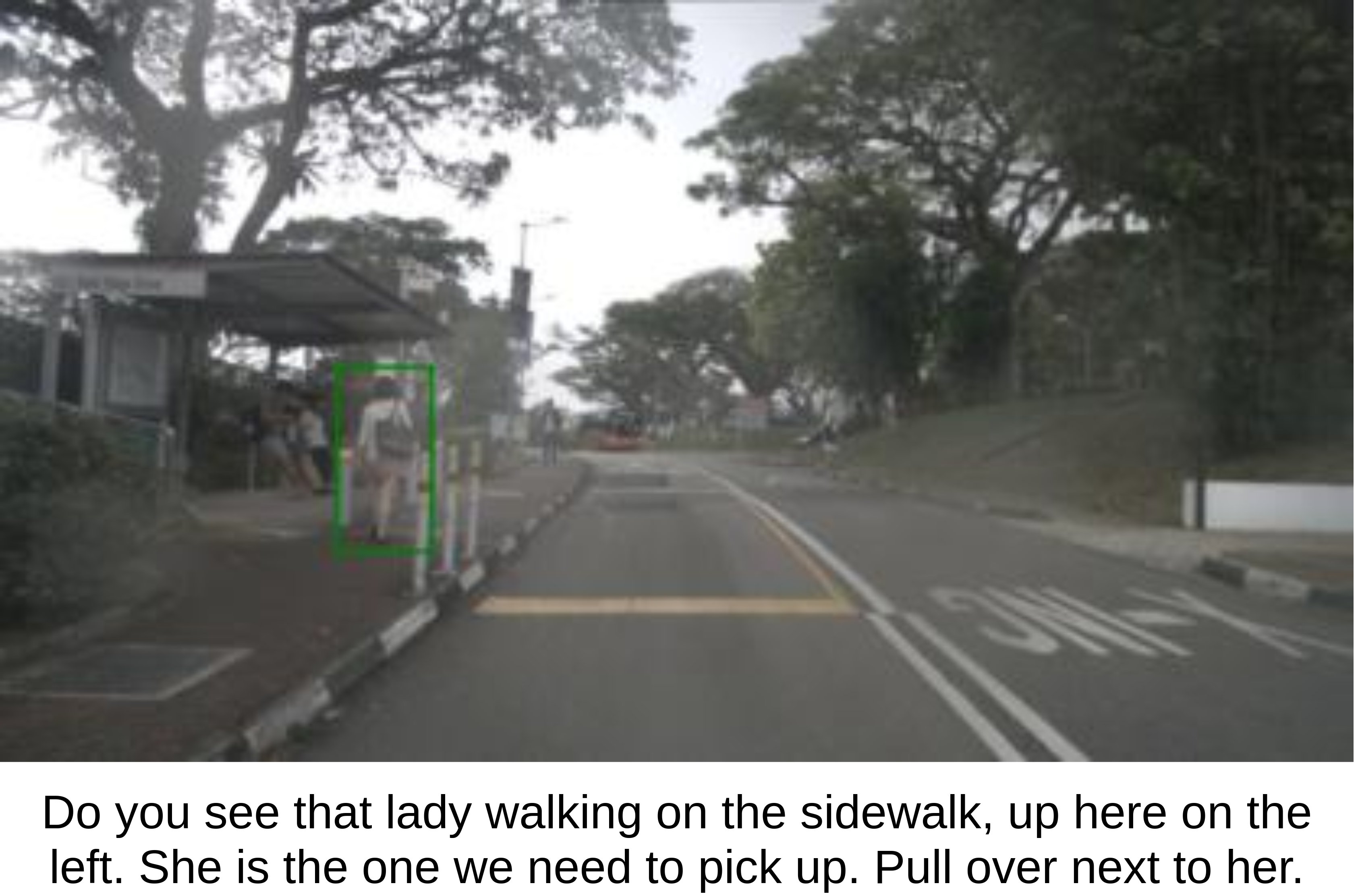}
\caption{Example of visual grounding task. Green box indicates the referred object.}
\label{fig:sample}
\end{figure}

In recent years, there have been many advances in tasks involving the joint processing of images and text. Researchers are working on various challenging problems like image captioning \cite{you2016image, anderson2018bottom, feng2019unsupervised}, visual question answering \cite{anderson2018bottom, jiang2020defense,yu2019deep}, text-conditioned image generation \cite{li2019controllable, zhang2017stackgan}, etc. In this work, we address the task of visual grounding \cite{deruyttere2020giving,sadhu2019zero} in which our goal is to train a model that can localize an image region based on a given natural language text query. The task of visual grounding can be useful in many practical applications. In Figure \ref{fig:sample},  we provide an example, in a self-driving car, the passenger can give a command to the car like “Do you see that lady walking on the sidewalk, up here on the left. She is the one we need to pick up. Pull over next to her”. Additional use cases can be found in embodied agents and human-computer interaction. The task of visual grounding is quite challenging as it requires joint reasoning over text and images. Consider the example shown in Figure \ref{fig:sample}: in order to correctly identify the women, a model first needs to understand the command properly and then locate the target object in the image.
\par
In this paper, we propose a novel architecture for visual grounding, namely, the AttnGrounder. At a high-level, visual grounding consists of two sub-tasks: object detection \cite{yolov1, faster_rcnn} and ranking detected objects. In the first task, the model identifies all the objects present in the image, and then it calculates a matching score between every object and given text query to rank different proposals. These two tasks are themselves quite challenging to solve, and the object detection task may become a bottleneck for the performance of the whole system. In contrast to the traditional two-stage \cite{yu2018mattnet, deruyttere2020giving} approaches our AttnGrounder directly operates on raw RGB images and text expressions. Our main goal is to combine the image and text features, and then directly predict bounding boxes from them. To this end, we use a YOLOv3 \cite{yolov3} backbone to generate visual features (Sec. \ref{sec:image_encoder}) and a Bi-LSTM to generates text features (Sec. \ref{sec:text_encoder}). For jointly reasoning over visual and text features, we use two sub-modules.: Text Feature Matrix (Sec. \ref{sec:text_feat_mtrx}) and Attention Map (Sec. \ref{sec:attention_map}). The text feature matrix calculates text representation for every image region by selecting important words for that particular region. Predicting a rough segmentation mask around the referred object can also help in predicting a bounding box. Thus, we make a rectangular mask around the target object using the ground truth coordinates, and the goal of our attention map module is to predict that mask. The attention map module is trained jointly with rest of the system and yields an auxiliary loss which can improve the localization ability of our model. Finally, we fuse all the features to obtain a multi-modal feature representation with which we predict the bounding box (Sec. \ref{sec:fusion}). In Section \ref{sec:related_work}, we give an overview of the related work. In Section \ref{sec:approach}, we provide a detailed description of our approach. In Section \ref{sec:exp}, we evaluate our model on the Talk2Car dataset \cite{talk2car}. Finally, we conclude our paper in Section \ref{sec:conclusion}.

\section{Related Work}
\label{sec:related_work}
In terms of the object detection subtask in visual grounding, there exist two lines of work in visual grounding: a one-stage approach \cite{yang2019fast, sadhu2019zero, chen2018real, luo2020multi} and a two-stage approach \cite{wang2019neighbourhood, cross_modal, liu2017referring, deruyttere2020giving}. In most of the previous work, researchers address the task of visual grounding with two-stage approaches. In the first stage, several object proposals are generated using an off-the-shelf object detection algorithm like Faster-RCNN \cite{faster_rcnn}, YOLO \cite{yolov1, yolov3}. In the second stage, the object proposals are ranked by calculating the matching score with the given text query.
\par
Several works \cite{cross_modal, wang2019neighbourhood} explore graph convolutional networks for learning the relationship between objects and text query. The work of \cite{cross_modal} focuses on improving the cross-modal relationship between objects and referring expressions by constructing a language-guided visual relation graph. They use a gated graph convolutional network to fuse information from different modalities. In \cite{wang2019neighbourhood}, authors also use graph networks for learning the relationship between neighboring objects. NMTree \cite{liu2019learning} constructs a language parsing tree with which they localize referred object by accumulating grounding confidence per node. MAttNet \cite{yu2018mattnet} involves a two-stage modular approach in which they use three modules for processing visual, language, location information, and a relationship module to understand the relationship between the output of these three modules. In \cite{liu2019improving}, the authors use attention to generate difficult examples by discarding the most important information from both text and image. A-ATT \cite{deng2018visual} constructs three modules to understand the image, objects, and query. They accumulate features from different modalities in a circular manner to guide the reasoning process in multiple steps.  MSRR \cite{deruyttere2020giving} also uses a multi-step reasoning procedure, in which a new matching score is calculated between every object and the language query in each step. After all the reasoning steps, the object that has the highest score is selected as the target object.
\par
In two-stage methods, the offline object detector may become a bottleneck as it may fail to provide good object proposals. For addressing this issue, several recent works explore a one-stage paradigm for visual grounding. One-stage methods fuse image-text features and then directly predict the bounding box for the referred object.
Recent works \cite{sadhu2019zero, yang2019fast} fuse visual-text features at every spatial location and predict adjustment in predefined anchor boxes to align the bounding boxes with the referred object. In \cite{chen2018real}, the authors introduce a guided attention module with which their model also predicts the center coordinates of the referred object along with the referred region. In our AttnGrounder, we predict a mask rather than the center coordinate which can help in better localization of image regions. Moreover, previous one-stage methods consider a single text representation for every location, but as we know, different regions correspond to different words, which is why we develop a text feature matrix where every location has its unique text representation. Additionally, one-stage methods are faster than two-stage methods as they don’t involve matching between various image regions and text queries.

\begin{figure}[t]
\centering
\includegraphics[height=3.6cm]{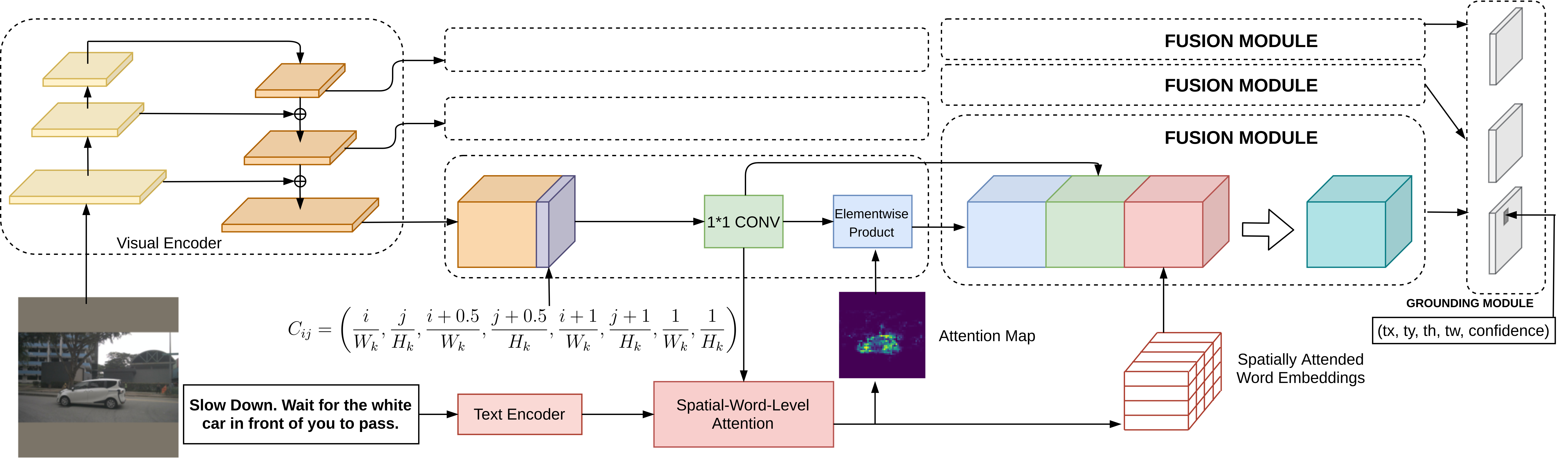}
\caption{Overview of AttnGrounder.}
\label{fig:complete_model}
\end{figure}

\section{Methodology}
\label{sec:approach}

In Figure \ref{fig:complete_model}, we provide an overview of our proposed visual grounding architecture. Our goal is to locate an object in an Image $I$ based on a given text query $T$. Our proposed model AttnGrounder consists of five sub-modules: image encoder, text encoder, visual attention module, fusion module, and a grounding module. The image encoder uses Darknet-53 \cite{yolov3} with pyramid structure \cite{lin2017feature} to extract visual features from the image $I$ in the form of grids $ \{G_k\}_{k=0}^{2} \in \mathbb{R}^{C_k\times H_k\times W_k}$ at three different spatial resolutions. To encode text features, we use a Bi-LSTM that can easily capture the long-range dependency in given query. We introduce a visual-text attention mechanism that attends to words $w_t$ in the text query $T$ at every spatial location in the grid $G_k$. This attention mechanism also generates an attention map $M$ for the referred object. Finally, we fuse the image and text feature using $1\times1$ convolution layers. For training, we use two loss functions: binary cross-entropy loss for training the attention map and YOLO's loss \cite{yolov1} function for complete end-to-end training. For our use, we modify the YOLO's loss function by replacing the last sigmiod unit with a softmax unit. 
\subsection{Image Encoder}
\label{sec:image_encoder}
We adopted Darknet-53 \cite{yolov3} with a pyramid network \cite{lin2017feature} structure for encoding visual features. Darknet-53 takes an image $I$ and produces feature grids $ \{G_k\}_{k=0}^{2} \in \mathbb{R}^{C_k\times H_k\times W_k}$ at three different spatial resolutions, where $C_k, H_k$ and $W_k$ are the number of channels, height and width of the grid at $k^{th}$ resolution. 
\\
Usually, referring expressions also contain position information about the referred object (eg. "park in front of the second vehicle on our \textit{right} side."). Visual features produced by Darknet lack such location information. Similar to \cite{yang2019fast}, we explicitly add location information by concatenating a vector $C_{ij} \in \mathbb{R}^{8}$ at every spatial location of the grid $G_{k}$,
\[C_{ij} =  \left( \frac{i}{W_k}, \frac{j}{H_k}, \frac{i + 0.5}{W_k},  \frac{j + 0.5}{H_k} , \frac{i + 1}{W_k}, \frac{j + 1}{H_k} , \frac{1}{W_k}, \frac{1}{H_k}\right) \]
where $i, j$ are row and column in the grid $G_k$ respectively. Then, we use a $1 \times 1$ convolution layer with batch normalization and ReLU unit to map these grid feature to a common semantic dimension $D$. Now, $G_{k} \in \mathbb{R}^{C_k\times H_k \times D}$. 
\subsection{Text Encoder}
\label{sec:text_encoder}
Our text encoder consists of an embedding layer and a Bi-LSTM layer. The text query $T$ of length $n$ is first converted to its embedding $Q = \{e_i\}_{i=0}^{n-1}$ using the embedding layer. After that, $Q$  is fed as an input to Bi-LSTM that generates two hidden states $[h^{right}_i, h^{left}_i]$ for every word embedding $e_i$. The forward and backward hidden states are concatenated to get $h_i \in \mathbb{R}^{2L}$, where $L$ is the number of hidden units in Bi-LSTM. Furthermore, we use a linear layer to project the text embedding into the common semantic space of dimension $D$. Embeddings for all $n$ words are stacked to get $Q^\prime \in \mathbb{R}^{n \times D}$.

\subsection{Visual-Text Attention}
\label{visual_text_attention}
\begin{figure}
\centering
\includegraphics[height=6.5cm]{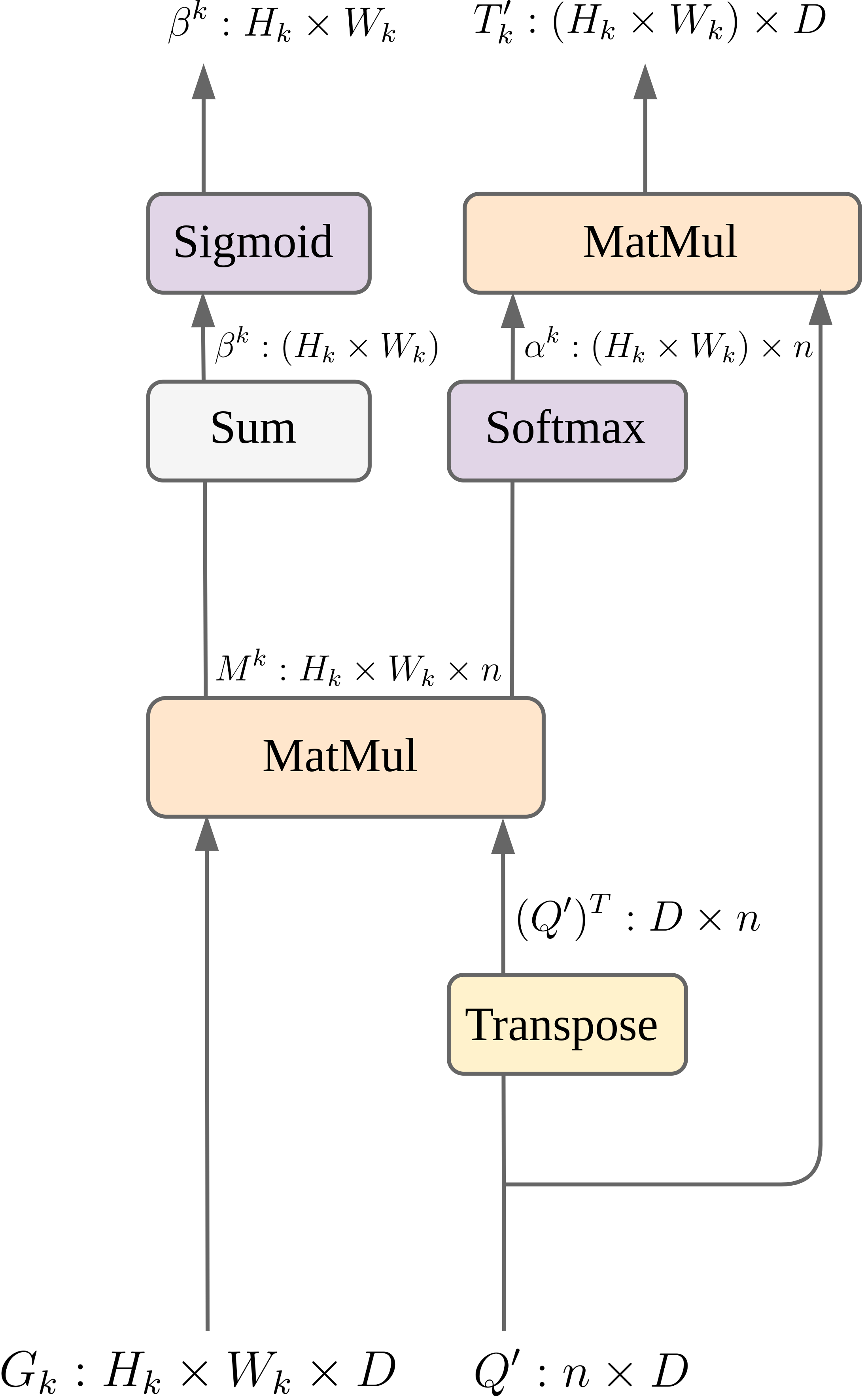}
\caption{Visual-Text Attention Module}

\label{fig:visual_text_attn}
\end{figure}

Every region in an image may correspond to different words in the given query. To better model the dependency between a word and a region, we present a visual-text attention module. Fig. \ref{fig:visual_text_attn} illustrates this module. This module gets a grid $G_k$ from the image encoder and text embeddings $Q^\prime$ from the text encoder. 
The grid $G_k \in \mathbb{R}^{H_k \times W_k \times D}$ has $H_k \times W_k$ spatial locations each of dimension $D$ and text embeddings $Q^\prime \in \mathbb{R}^{n\times D}$ have $n$ word features each of dimension $D$. To measure the matching score between a location $g_i \in \mathbb{R}^{D}$ and word $w_j \in \mathbb{R}^{D}$ in the query, we calculate the dot product between them. We multiply $G_k$ with $Q^\prime$, which generates a matrix $M^{k} = G_{k}\cdot (Q^\prime)^{T}$, where $M_{ij}^{k}$ represents the matching score between grid location $i$ and word $j$. The matrix $M^{k} \in \mathbb{R}^{(H_k \times W_k) \times n}$, where $H_k$ and $W_k$ are the height and width of the grid at $k^{th}$ resolution generated by our image encoder. This matrix $M^k$ is used in two ways described next.  

\subsubsection{Text Feature Matrix}
\label{sec:text_feat_mtrx}
For every location in the grid $G_k$, we can represent it's text features by summing up word features from the matrix $Q^\prime \in \mathbb{R}^{n\times D}$. We normalize the matrix $M^k$ using the softmax function to generate a word-level attention matrix $\alpha^k \in \mathbb{R}^{(H_k \times W_k) \times n}$, where $\alpha_{ij}^k$ denotes the correlation between region $i$ in the grid $G_k$ and word $j$ in the text query $T$. 
\begin{equation}
    \alpha_{ij}^k = \frac{\exp(M_{ij}^k)}{\sum_{j= 0}^{n-1} \exp(M_{ij}^k)} 
\end{equation}

To obtain text features at every location, we multiply $\alpha^k$ and $Q^\prime$, which generates text features matrix $ T^{\prime}_k \in \mathbb{R}^{(H_k \times W_k) \times D}$, denoted as $T^{\prime}_{k}= \alpha^k \cdot Q^\prime$. Every row in the matrix $T^{\prime}_k$ represents text features, which are effectively derived from the word features weighted by the correlation between words and corresponding image region. Thus, every region has its unique text representation.  

\subsubsection{Attention Map}
\label{sec:attention_map}
In visual grounding, a rectangular mask around the referred object can be predicted by just using the provided ground truth bounding boxes. Our attention map module aims to make a rectangular mask around the referred object, which is an auxiliary task that helps improve the overall performance. By learning to make a mask around the object, the model can indirectly learn to locate objects based on the given query and image. We make use of the matrix $M^{k} \in \mathbb{R}^{(H_k \times W_k) \times n}$, which already contains the matching score between every image region and every word in the text query. We sum the values in every row of $M^k$ to get a column matrix $\beta^{k} \in \mathbb{R}^{(H_k \times W_k)}$ 
\begin{equation}
    \beta^k_i = \sum_{j = 0}^{n-1} M_{ij}^{k}
\end{equation}
$\beta^k_i$ provides the aggregated matching score for an image region $i$ and the given text query $T$. Now, the image regions matching with the given query will have a high matching score and vice-versa. To create a mask, we feed $\beta^k$ through a sigmoid unit, which maps the values of $\beta^k$ between 0 and 1. Therefore, $\beta^k_i$  can be interpreted as the probability of having the referred object in the region $i$.
\par
\begin{figure}[t]
\centering
\includegraphics[height=4.3cm]{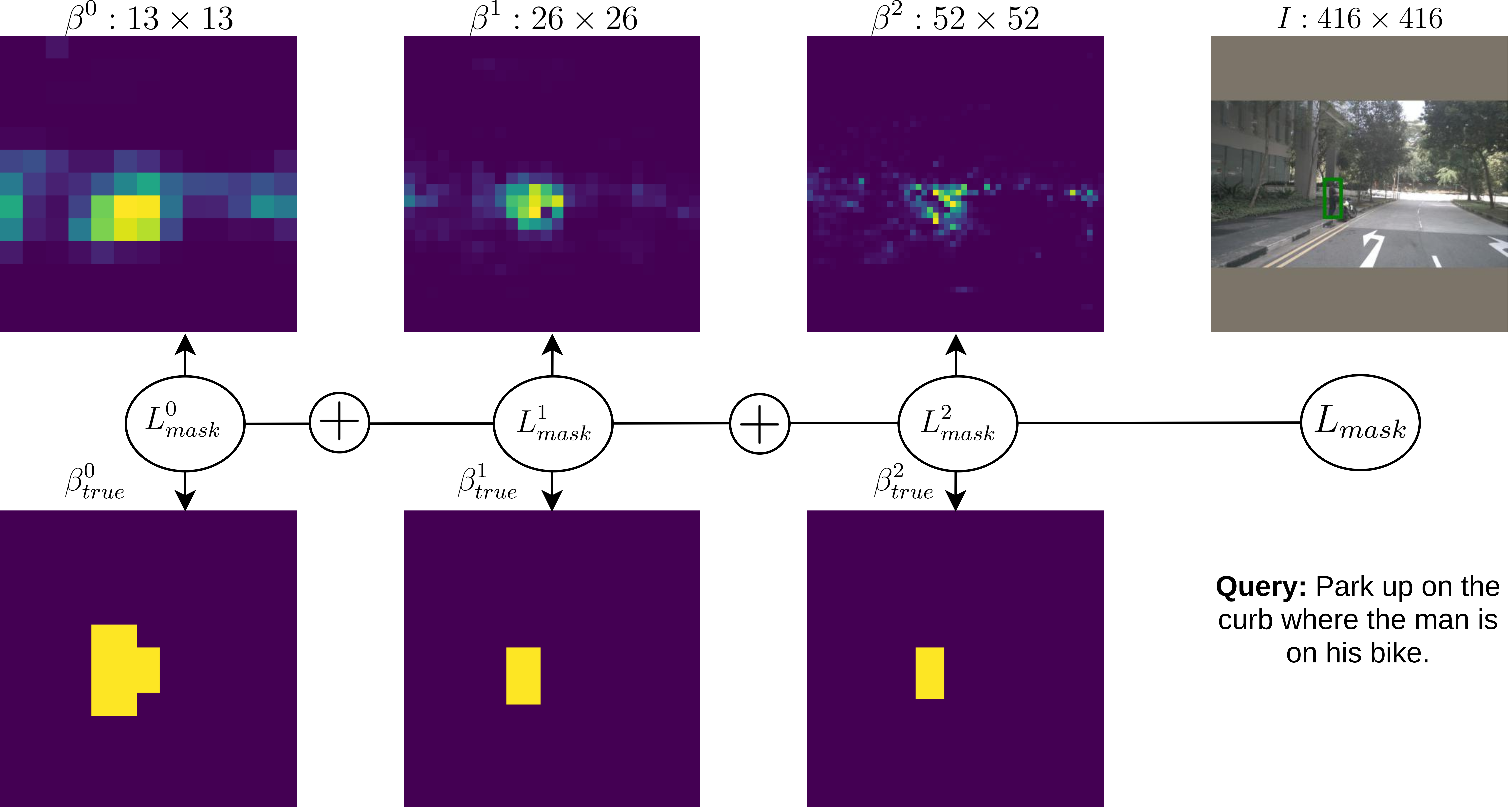}
\caption{The mask loss is calculated for every spatial resolution using binary cross-entropy loss and summed to get the final mask loss. First row shows attention map produced by our AttnGrounder at three different resolutions and second row shows ground truth mask used for training.}
\label{fig:attention_loss}
\end{figure}

For training, we generate a rectangular mask using the ground truth coordinates. The ground truth mask has the value of 1 inside the rectangular region and 0 elsewhere. Masks are created for all three resolutions and training is done using binary cross-entropy loss (See Figure \ref{fig:attention_loss}). At the $k^{th}$ resolution, given the ground truth mask $\beta^{k}_{true} \in \mathbb{R}^{H_k \times W_k} $ and the predicted mask $\beta^{k} \in \mathbb{R}^{H_k \times W_k}$; the mask loss is calculated as 
\begin{equation}
    L^{k}_{mask} = - \frac{1}{H_k \times W_k}\sum_{i = 0}^{(H_k \times W_k) - 1} (\beta^{k}_{i, true}\log(\beta^{k}_{i}) + (1 - \beta^{k}_{i, true})\log(1 -\beta^{k}_{i}) )
\end{equation}
The $L_{mask}$ is obtained by summing up the $L_{mask}^k$ terms calculated for all the three resolutions. 
\begin{equation}
    L_{mask} = \sum_{k = 0}^{2} L_{mask}^k
\end{equation}
Using $L_{mask}$ as an auxiliary loss can help in embedding visual and text features in the common semantic space $D$, by encouraging similar visual and text features to be nearby and vice-versa. Furthermore, it can also help discriminate between the referred region and the others.   
\par
We also generate attended visual features $V_k \in \mathbb{R}^{H_k \times W_k \times D}$ by an element-wise product between the attention map $\beta^k$ and grid $G_k$, $V_k = \beta^k \odot G_k$ ($\odot$ denotes element-wise product). In $V_k$, every spatial location is scaled by its importance determined by the attention mask $\beta^k$. Thus, visual features of regions where target object may be present are enhanced and visual features of other redundant regions are reduced.

\subsection{Fusion Module}
\label{sec:fusion}
For fusing image and text features, we concatenate visual feature grid $G_k$, text feature matrix $T^{\prime}_k$ and attended visual features $V_k$ along the channel dimension. This generates a matrix $F_k \in \mathbb{R}^{H_k \times W_k \times 3D}$. All the features are $l_2$ normalized before concatenation and fusion is done for all the three spatial resolutions generated by our image encoder. After that, we use a $1 \times 1$ convolution layer to fuse these visual and text features. This $1 \times 1$ convolution layer maps these fused features in a new semantic space of dimension $D^\prime$. After fusion, we have three feature matrices $\{ F_k\}_{k=0}^{2} \in \mathbb{R}^{H_k \times W_k \times D^\prime}$.

\subsection{Grounding Module}
\label{sec:grounding_module}
The grounding module aims to ground the text query onto an image region. This module takes the fused feature vector $F_k$ as input and generates a bounding box prediction. We follow \cite{yang2019fast} for designing this module and similar to \cite{yang2019fast} we replace YOLO's \cite{yolov1} output sigmoid layer with softmax layer.
\par
For every spatial location, YOLOv3 centres three different anchor boxes and for every spatial resolution, YOLOv3 uses different set of anchor boxes. In total, for three different spatial resolutions we have $3 \times 3 = 9$ different predefined anchor boxes. We denote the total number of anchor box predictions made by our model with $m$, where $m = \sum_{k = 0}^{2} H_k\times W_k \times 3$. For every anchor box, YOLOv3 predicts changes required in location and size to fit that particular anchor box around the target object. Specifically, YOLOv3 uses two branches: the first branch predicts the shift in centre, height, width of the predefined anchor box and the second one uses a sigmoid layer to predict the confidence (on the shifted box) of being the target box. As we need only one bounding box prediction for grounding, we replace the last sigmoid layer with softmax layer which forces to select one box out of the $m$ boxes as prediction. Accordingly, the loss function for confidence is also changed to a cross-entropy loss between the predicted confidence (softmax version) and a one-hot vector with a 1 entry corresponding to the anchor box that has the highest intersection over union with the ground truth box. We refer the readers to \cite{yolov1, yolov3} for more details and abstract away the YOLO's loss function as $L_{yolo}$. Thus, total loss becomes 
\begin{equation}
    L = L_{yolo} + \lambda L_{mask}
\end{equation}
where $\lambda$ is a hyper-parameter which scales the loss $L_{mask}$.

\begin{figure}[t]
\centering
\includegraphics[height=4cm]{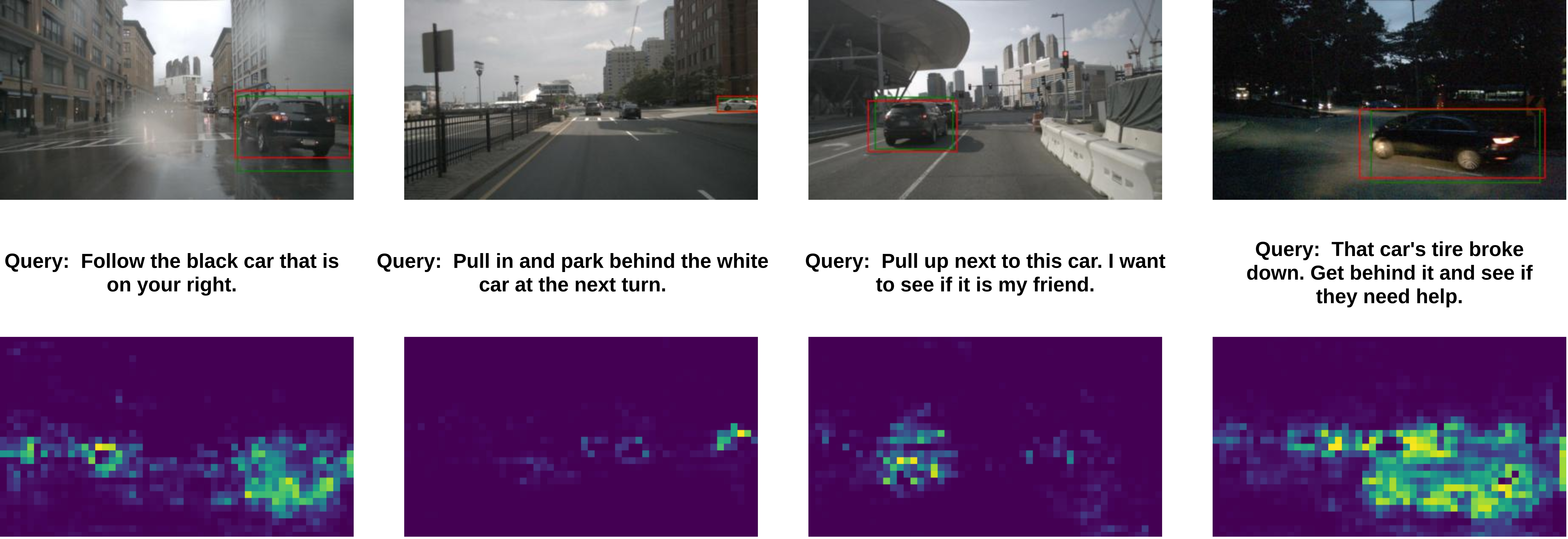}
\caption{Examples of results obtained using our AttnGrounder. The predicted bounding boxes are shown in red and ground truth bounding boxes are shown in green. The second row shows the attention maps generated by our model.}
\label{fig:examples}
\end{figure}

\section{Experiments and Results}
\label{sec:exp}
In this section, we provide details of our experiments. We evaluate our approach on Talk2Car dataset \cite{talk2car} and compare with five baselines. 
\subsection{Dataset}
The Talk2Car dataset is based on nuScenes dataset \cite{nuScenes} and contains 11,959 referring expressions for 9,217 images. This dataset contains images of a city-like environment where a self-driving car is driving. Talk2Car dataset is a multi-modal dataset that consists of various sensor modalities such as semantic maps, LIDAR, 360-degree RGB images, etc. However, we only use the RGB images for training our AttnGrounder. RGB images contained in Talk2Car dataset are taken in different weather conditions (sunny, rainy) and time of day (day, night) which makes it even more challenging.  Additional challenges that the Talk2Car dataset presents are the ambiguity between the objects belonging to the same class, far away object localization, long text queries, etc. In this dataset, the size of images is $900 \times 1600$ and on average, the referring expressions contain 11 words. The dataset contains 8349, 1163 and 2447 images for training, validation and testing respectively. Figure \ref{fig:sample} shows an example of Talk2Car dataset.

\subsection{Training Details}

Our image encoder backbone i.e. Darknet-53\cite{yolov3} is pre-trained on COCO \cite{lin2014microsoft} object objection task. We use 300 dimensional GloVe \cite{pennington2014glove} embeddings for initializing word vectors. Words for which the GloVe embeddings were not available were initialized randomly. We resize the images to $416 \times 416$ and preserve the original aspect ratio. We resize the longer edge to 416 and pad the shorter edge with the mean pixel value. We also add some data augmentations i.e., horizontal flips, changing saturation and intensity, random affine transformation. The anchors used in our grounding module are (10,13), (16,30), (33,23), (30,61), (62,45), (59,119), (116,90), (156,198), (373,326). We use $\lambda = 0.1$ (defined in Sec. \ref{sec:grounding_module}) for training.
We train our model using Adam \cite{adam} optimizer with an initial learning rate of $10^{-4}$ and polynomial learning rate scheduler is used with a power of 1. 
For the pre-trained Darknet-53 portion, we keep a lower initial learning rate of $10^{-3}$. We keep a batch size of 14 for training our AttnGrounder. 

\begin{table}[t]
    \centering
    \caption{Comparison with state-of-the-art methods on Talk2Car testset}
    \label{table:results}
    \begin{tabular}{ |p{3cm}|P{2cm}|P{2cm}|P{2cm}| }
    \hline
        Method                              & $AP_{50}$ Score & Time (ms) & Params (M)\\
        \hline
        STACK \cite{hu2018explainable}                              & 33.71 &52  & \textbf{35.2} \\
        SCRC \cite{hu2016natural}                                & 43.80 &208 & 52.47\\
        A-ATT \cite{deng2018visual}                               & 45.12 &180 & 160.31\\
        MAC \cite{hudson2018compositional}                                & 50.51 &51 & 41.59\\
        MSRR \cite{deruyttere2020giving}                            & 60.04 &270.5 & 62.25\\
        \hline
        \textbf{AttnGrounder}        & \textbf{63.30} & \textbf{25.5} & 75.84\\
    \hline
    \end{tabular}
    \label{tab:my_label}
\end{table}

\subsection{Comparison Metrics}
We evaluate our approach on three different metrics. First one is $AP_{50}$ score, which is defined as the percentage of predicted bounding boxes that have an \textit{Intersection Over Union} of more than 0.5 with the ground truth bounding boxes. The second one is \textit{inference time} and third is \textit{number of parameters} in a method.

\subsection{Results}
We compare our AttnGrounder with four baselines and state-of-the-art model MSRR \cite{deruyttere2020giving}, results on the test set of Talk2Car dataset are shown in Table \ref{table:results}. AttnGrounder outperforms all the baselines and improves state-of-the-art by 3.26\% in terms of $AP_{50}$ score. Thanks to our one-stage approach, we also achieve lowest inference time among all the baseline methods. We provide examples of prediction made by our method in Figure. \ref{fig:examples} along with the visualization of attention map generated by visual-text attention module (Sec. \ref{visual_text_attention}).

\section{Conclusions}
\label{sec:conclusion}
We propose AttnGrounder, a single-stage end-to-end trainable visual grounding model. Our AttnGrounder is fast compared to two-stage approaches as it does not involve matching various region proposals and text queries or multi-step reasoning. We combine visual features extracted from YOLOv3 and text features extracted from Bi-LSTM to obtain a multi-modal feature representation with which we directly predict the bounding box for the target object. Our AttnGrounder also generates an attention map that localizes the potential spatial locations where the referred object may be present. This attention map is trained as an auxiliary task which helps improve the overall performance of our model. Finally, we evaluate our proposed method on the Talk2car dataset and show that it outperforms all the baseline methods.
\section*{Acknowledgements} 
The author wants to thank Thierry Deruyttere, Dusan Grujicic, and Monika Jyotiyana for their feedback on an early draft of this paper. The author is grateful to Thierry Deruyttere for his guidance and constant support.

\clearpage
%
%
\bibliographystyle{splncs04}
\bibliography{refer.bib}
\end{document}